\documentclass{research4cacm}
\begin{document}
%
\newtheorem{conjecture}{Conjecture}

\title{Deep Learning Works in Practice. \\ 
But Does it Work in Theory?
%
}
%
%
%
%
%
\numberofauthors{2} 
%
\author{
%
%
\alignauthor
L\^e Nguy\^en Hoang\\
       \affaddr{EPFL}\\
       \email{le.hoang@epfl.ch}
\alignauthor
Rachid Guerraoui\\
       \affaddr{EPFL}\\
       \email{rachid.guerraoui@epfl.ch}
}

\maketitle
\begin{abstract}
{\it Deep learning}  relies on a very specific kind of neural networks: those superposing {\it several} neural layers.
In the last few years, deep learning achieved major breakthroughs in many tasks such as image analysis, speech recognition, natural language processing, and so on. Yet, there is no theoretical explanation of this success. In particular, it is not clear why the {\it deeper}  the network, the {\it better}  it actually performs. 

We argue that the explanation is intimately connected to a key feature of the {\it data} collected from our surrounding universe to feed the machine learning algorithms: {\it large non-parallelizable logical depth}. Roughly speaking, we conjecture that 
the shortest computational descriptions of the universe are algorithms with inherently large computation times, even when a large number of computers are available for parallelization. Interestingly, this conjecture, combined with the  folklore conjecture in theoretical computer science that $ P \neq  NC$,  explains the success of deep learning. 

\end{abstract}

\section*{Turing Test and Learning Machines}

In 1950, Alan Turing \cite{turing1950} boldly conjectured that passing the now-called {\it Turing test} would require {\it machine learning}. His amazing insight was essentially the following. The human brain has around $10^{15}$ synapses and $10^9$ of these synapses are likely critical to the kind of natural language processing needed to pass the Turing test. Turing went on regarding this quantity as the minimal amount of bits  needed for any algorithm to pass the test. In modern terms, we would say that the {\it Kolmogorov complexity}  \cite{kolmogorov1963} of the Turing test is likely to be of the order of $10^9$ bits. This corresponds to saying that no shorter algorithm can solve this problem.


Unfortunately, writing quality source codes that are $10^9$ bits long is extremely challenging, tedious, time-consuming and prone to errors, even for large teams of top software developers. Recall that there are around $2^{10^9}$ such source codes, which is much larger than the number of particles in the universe. Back in 1950, Turing foresaw that, around 2000, computer science would undergo a revolution, as it would rely more and more on machine-written rather than hand-written algorithms. Machine learning (or {\it  learning machines}, as Turing called it) would thus correspond to the superhuman ability of computers to explore the space of algorithms whose Kolmogorov complexity exceeds the gigabyte. Of course, the key to this new capability of computers is their superhuman computational power, as well as the availability of huge amounts of data to guide them in their exploration of $10^9$-bit-long source codes. 

Now, Turing did not himself specify what machine learning algorithm to use.
Few years later, Solomonoff \cite{solomonoff1964a,solomonoff1964b} proposed applying Bayes' rule to all terminating Turing machines. Denoting $\mathcal T$ the set of such machines and $D$ the observed data, Solomonoff's machine learning algorithm, called {\it Solomonoff's induction}, consists of computing the posterior credence $\mathbb P[T|D]$ of a terminating Turing machine $T$ given data $D$ by
$$\mathbb P[T|D] = \frac{\mathbb P[D|T] \mathbb P[T]}{\sum_{S \in \mathcal T} \mathbb P[D|S] \mathbb P[S]},$$
where $\mathbb P[D|T]$ is the probability that the Turing machine $T$ assigns to data $D$ and $\mathbb P[T]$ is the prior probability of the Turing machine $T$ which, for the sum of priors to equal 1, will typically have to be exponentially small in the size of the description of $T$.
Interestingly, Solomonoff proved two things:

\begin{enumerate}

\item Solomonoff proved that his very general approach to machine learning was {\it  complete}, in the sense that it provably determined the best-possible predictive theory from an amount of data whose size is (roughly) the Kolmogorov complexity of the data itself.

\item Solomonoff also proved however that his induction was  {\it  incomputable}. In short, this is because the space of all terminating Turing machines that Solomonoff proposed to explore is {\it  ill-behaved}. Indeed, as Turing \cite{turing1937} showed in 1936 through the infamous halting problem, this space is not a recursively enumerable set. (Besides, Solomonoff's use of Bayes' rule is way too computationally costly to be applied in practice.)

\end{enumerate}

This naturally led to rather focus on restricted {\it  better-behaved} computational models, e.g. {\it linear classifiers}, whose exploration is facilitated by some easy-to-compute learning rule, such as {\it (stochastic) gradient descent}. This allows to automate the data-driven exploration of a {\it restricted} subspace of large-Kolmogorov-complexity algorithms, in a way that humans cannot match. 
This is arguably why {\it machine learning works in theory}.
 What is less clear is why {\it deep learning} is particularly effective. Before discussing this, let us first recall what {\it deep} actually means which, in turn, requires to recall some basic elements of neural networks.

\section*{Neural Networks: the Deeper the Better}

Neural networks are often regarded as the most promising restricted computational model for machine learning. 
 Neurons in such a network can be thought of as elementary processing units, while the directed edges of the network can be thought of as communication channels between processing units. A crucial feature of neurons, which we shall get back to below, is that they perform fast nonlinear operations. Neurons typically compute a linear combination of incoming signals, and then apply a sigmoid function or a piecewise linear transformation to the linear combination.\footnote{Note that the coefficients of the linear combination are usually associated to the communication channels, also known as synapses or edges. These two ways of defining neural networks are however equivalent. Yet, here, we shall clearly distinguish communication steps from computation steps.} But neurons may as well compute other fast nonlinear operations, e.g. softmax, max-pooling or energy-based sampling.

One popular neural network architecture is the  {\it  feed-forward} one \cite{rumelhart1988}. In this particular network, neurons are organized in layers. The outer layer is fed with raw data. Neurons of this outer layer then communicate their data to some (or sometimes all) of the neurons of the first hidden layer. These neurons perform their nonlinear transformation of their inputs, then communicate their results to neurons of the second layer, and so on. In this architecture, the number of neurons per layer is commonly known as the {\it width} of the neural network, while the number of layers is called its {\it depth}. {\it Deep learning} roughly boils down to favoring depth over width.\footnote{It is not clear though how to relevantly define depth for recurrent neural network.} This is in contrast with what is sometimes called {\it shallow learning}.

\begin{figure}
\centering
\includegraphics[width=.45\textwidth]{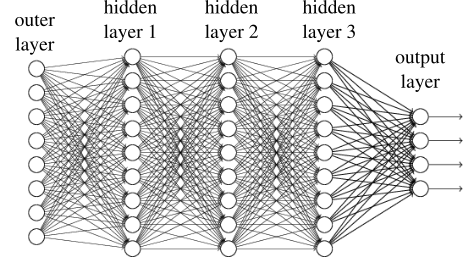}
\caption{A neural network with 3 hidden layers.}
\label{fig:neuralnet}
\end{figure}


An essential feature of neural networks is the ease with which they can be adjusted to data, usually through the (stochastic) gradient descent of a data-driven loss function combined with a {\it back-propagation} algorithm \cite{rumelhart1988}. Such a learning scheme typically corresponds to adapting the importance each neuron  gives to its different input signals in order to better fit the data, or even the effect of an input on the output.


Over the last few years, deep learning has allowed monumental breakthroughs in a huge number of areas, including image analysis, speech recognition, natural language processing, car driving, winning the game of Go, music composition, painting drawing, doodling and so on. We refer to \cite{lecun2015} for a more thorough list of deep learning success stories. The repeated and surprising successes of deep learning have convinced most practitioners of the fact that {\it deep learning works in practice}. \footnote{Given sometimes huge computation power: deep learning for image recognition for instance typically requires billions of images, whose mere processing was beyond the reach of the fastest computers a few decades ago. In fact, deep learning now typically requires GPU hardware.}



However, and as we pointed out earlier, while Turing anticipated the success of machine learning long ago, the success of deep learning, as opposed to other machine learning schemes,  remains a mystery. Many researchers often assert that no one quite really understands why deep learning performs so well. Yann LeCun, one of the main pioneers of deep learning, talks about the "unreasonable effectiveness of deep learning"\cite{lecun2014}.


\begin{figure}
\centering
\includegraphics[width=.45\textwidth]{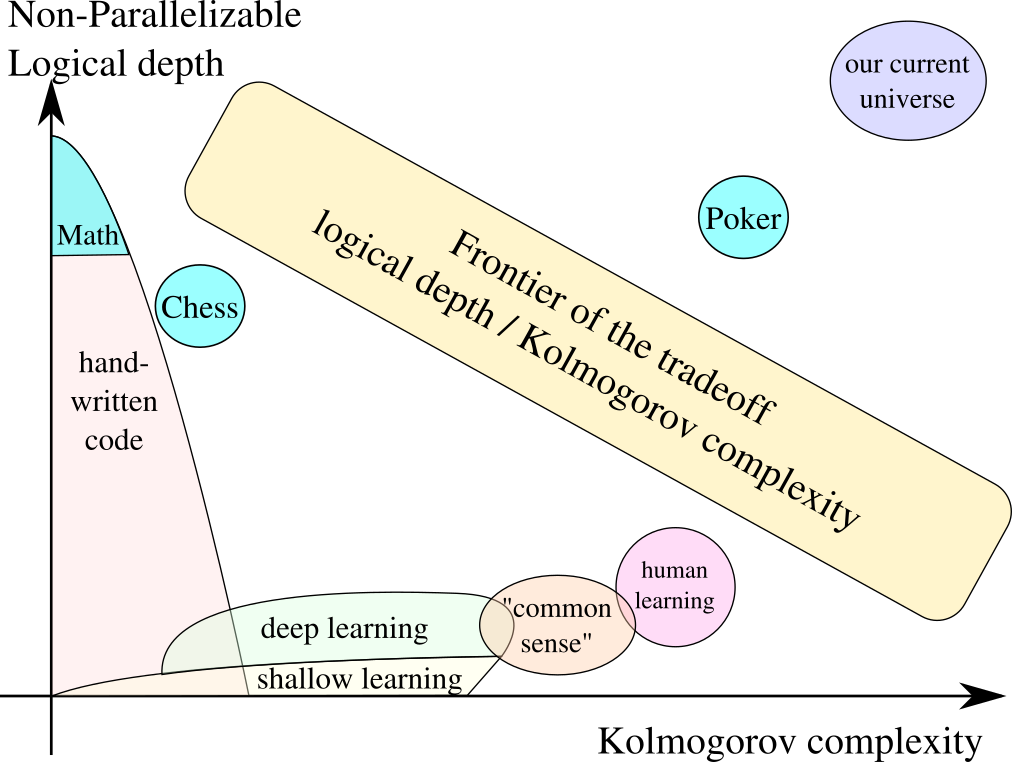}
\caption{Space of algorithms and problems.}
\label{fig:tradeoff}
\end{figure}

\section*{Conjectures}

We present below three conjectures. 
The first can be viewed as a rephrasing of Turing's argument in modern terminology. As Turing would likely argue, it explains why machine learning often works better than human-written codes.

\begin{conjecture}
Most of the data from the current state of our universe and most of the problems we aim to solve with these data, as well as any good approximations of these data and problems, have a Kolmogorov complexity larger than $10^9$ bits.
\end{conjecture}

Our second and main conjecture essentially says that the structures  we observe in our daily lives have an inherent  {\it apparent complexity}, or {\it interestingness}. {\it Apparent complexity} can be precisely captured by the notion of {\it logical depth}, introduced by Bennett \cite{bennett1995}. More specifically,  the logical depth of a datum, e.g. a video or a data set of images, is the computation time of the shortest algorithm that outputs this datum.\footnote{To avoid degeneracy, this definition is sometimes replaced by the smallest computation time of algorithms whose length is at most the Kolmogorov complexity plus a small fixed constant.} We argue that data from our daily lives typically feature large logical depth. In other words, our data can usually be remarkably well compressed, and the decompression of the optimal data compression often requires large computational power. 

\begin{conjecture}
Most of the data from the current state of our universe and most of the problems we aim to solve with these data, as well as any good approximations of these data and problems, have a large non-parallelizable logical depth. 
\end{conjecture}

We will discuss below the importance of {\it non-parallelizability} and give several examples arguing for {\it logical depth}. Before doing so however, let us first better explain what the conjectures actually imply. 
Essentially, the two conjectures above say that the full description of our universe requires both large Kolmogorov complexity {\it and} large non-parallelizable logical depth. In fact, it seems that many classical problems can be located somewhere in a diagram of two axes that correspond to these two distinct measures of computational complexity. The same seems to hold for  machine learning algorithms as well.  Figure \ref{fig:tradeoff} depicts the relative complexity of different problems and algorithms. 

Now, in order to understand the relation with deep learning, it is important to observe that, strictly speaking, a deep neural network does not perform more computation steps than a shallow one. 
Indeed, each neuron performs a computation step. In this sense, the number of computation steps of a neural network corresponds to the number of neurons it features. The relevant concept of depth of a problem, as highlighed in our main conjecture, is precisely that of {\it non-parallelizable logical depth}. 

Non-parallelizable logical depth is intimately connected to the fundamental open question in theoretical computer science {\it  P versus NC}.  This question asks whether problems that can be solved in polynomial time on a Turing machine can be solved in polylogarithmic time on a polynomial-size logic gate circuits. It is widely believed that this is not the case: some polynomial-time problems are fundamentally non-parallelizable. This intuition seems precisely to be corroborated by the success of deep learning over shallow learning, as deep learning seems able to compute functions of large non-parallelizable logical depth that highly parallelized shallow neural networks cannot. In fact, this is our third conjecture.


\begin{conjecture}
At equivalent Kolmogorov complexity, deeper neural networks compute functions with larger non-parallelizable logical depth.
\end{conjecture}

Note that the $P \neq NC$ conjecture would only represent an asymptotic version of what is needed for our third conjecture. Having said this, given that the size of each input of neural networks is often less than a gigabit, a logarithmic-time function of such inputs would typically terminate in at most $\log_2(10^9) \approx 30$ steps, which could be computed by a neural network of depth 30. Thus, if $P = NC$, one would expect that deep learning with a lot more than 30 layers does not significantly outperform neural networks with 30 layers. However, current state-of-the-art deep learning algorithms can have "over 1200 layers and still yield meaningful improvements" \cite{huang2016}. This seems like a strong evidence for $P \neq NC$.


Let us recapitulate. On the one hand, hand-written codes have successfully determined solutions to large logical depth problems, like playing chess, but they are limited in their Kolmogorov complexity. On the other hand, shallow machine learning allows a better exploration of a subspace of large Kolmogorov complexity algorithms. However, shallow machine learning excludes all algorithms whose (parallelized) computation times exceed a few (non-parallelized) computation steps. Considering our three conjectures, both of these approaches inevitably underperform, because our current state of the universe and many classical problems seem to precisely require algorithms of both (a) large Kolmogorov complexity {\it as well as} (b) large non-parallelizable logical depth. Deep learning is the current state-of-the-art approach to efficiently explore a space of algorithms with both properties. We argue that this is why {\it deep learning works in theory}.

\section*{Corroborating the Main Conjecture}

To corroborate our main conjecture,  saying that the data used to feed our algorithms have an {\it apparent complexity}, consider the following examples. 

\begin{figure}
\centering
\includegraphics[width=.45\textwidth]{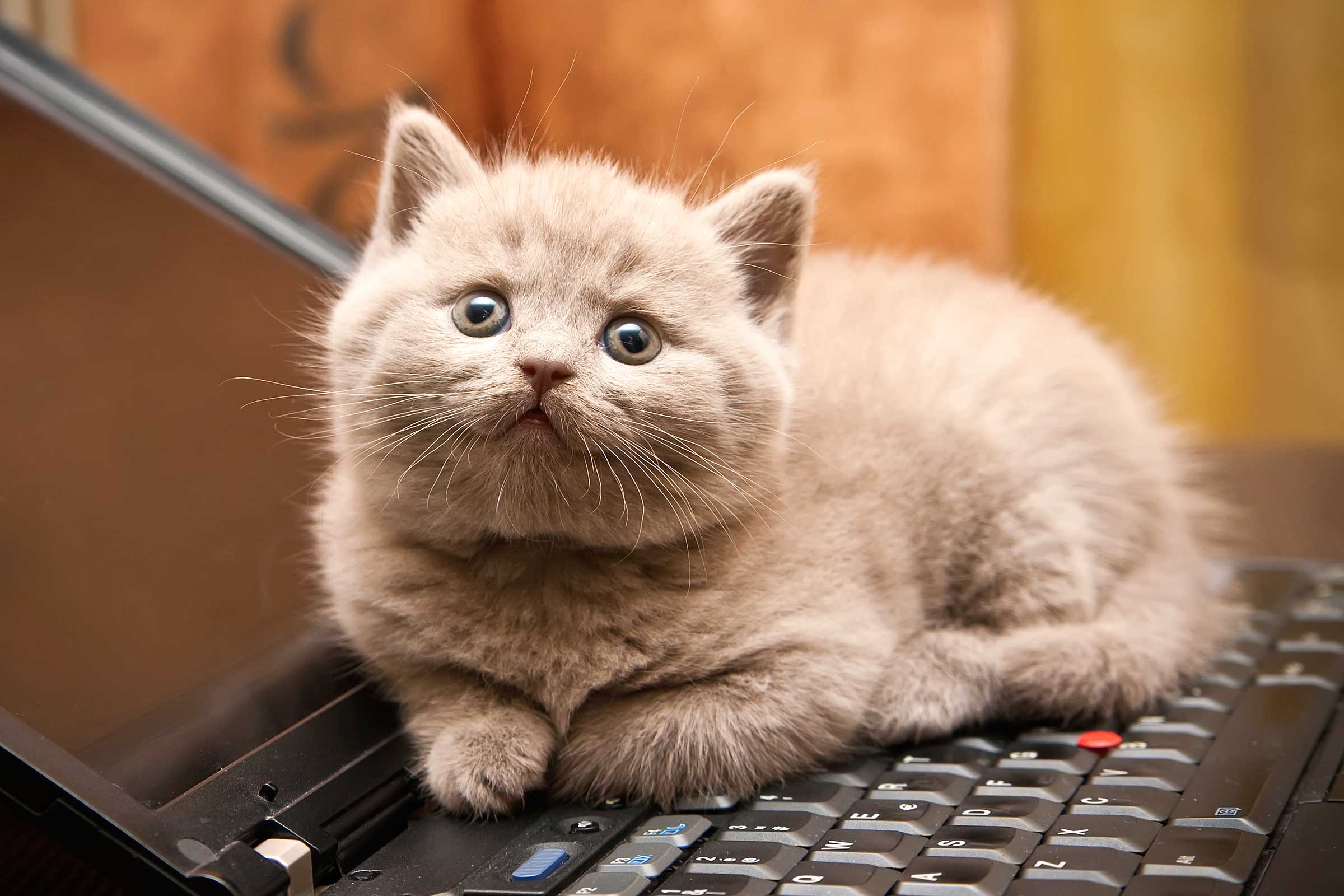}
\caption{A beige, puffy and sad-looking cat sitting on a laptop keyboard.}
\label{fig:cat}
\end{figure}

\begin{enumerate}

\item  Think of the way we, as humans, would describe many of the pictures of the web. Perhaps we would say that, on some image, we see a cat sitting on a laptop keyboard, that the cat is beige and puffy and that it looks sad. Somehow, however incomplete, this amazingly short description of the image successfully describes much of the information carried through the pixel luminosities and colours of a potentially high-definition image. Clearly, some thinking seems required to decompress  and to visualize the scene with this information. We invite the reader to do this effort, before comparing what she imagined with Figure \ref{fig:cat}.

\item  While sound could a priori be any kind of time series, most of our music is actually highly codified. Music is played by a small number of instruments with characteristic timber. Instruments play a handful of possibles pitches at a handful of possible rhythms, and even the combination of pitches and rhythms is often restricted by the choice of music genre. One can thus describe music in a very efficient manner, typically by a ZIP compression of music sheets. Yet,  deriving the actual music from its efficient description may require a lot of computation.

\item  Morphogenesis is a wonderfully sophisticated biological computation that has captivated Turing \cite{turing1952}. It is the ability of a single egg to morph into an actual living organism. Organisms can thereby be regarded as structures of large logical depth. Indeed, most of the information about this structure can be found in a DNA whose size is usually of a few (probably compressable) megabits. This corresponds to a reasonably small Kolmogorov complexity. However, deriving the future structure of the living organisms from its genome is likely to necessarily require a huge amount of computation steps. Indeed, gestation typically lasts months.

\item  Fractals are well-known for being computationally demanding. Indeed, fractals can often be obtained by a very simple procedure being recursively repeated over and over, as in the case of the Romanesco cabbage (see Figure \ref{fig:romanesco}). This typically corresponds to a low Kolmogorov complexity, but a large non-parallelizable computational depth. Intriguingly, simulations by \cite{poole2016} show that random deep neural networks compute more fractal-like structures than shallow ones do. 

\begin{figure}
\centering
\includegraphics[width=.45\textwidth]{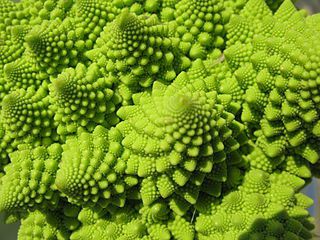}
\caption{Romanesco cabbage.}
\label{fig:romanesco}
\end{figure}

\item Non-parallelizable logical depth allows to better understand why many classical automata such as Wolfram's rule 30, Langton's ant or Conway's game of life are typically known to feature chaotic unpredictable phenomena, to the point where their unpredictability is sometimes used for pseudo-random number generation. This may seem somewhat contradictory with the fact that these automata are in fact deterministic and computable. However, we argue that what is actually meant is that such phenomena are unpredictable by our brains (or our machine learning models), because the logical depth of the phenomena far exceeds the depth of our brains (or even of our current deepest artificial neural networks).

\item There is a remarkable parallel between the success of deep learning and what Wigner \cite{wigner1960} famously dubbed "the unreasonable effectiveness of mathematics in the natural sciences". A common denominator to both  approaches to describing our world is the prevalence of depth. Indeed, mathematics can be regarded as the pinnacle of what humans have produced in terms of logical depth. Even though mathematical textbooks rarely exceed a thousand pages, their understanding often require years of study and a huge amount of cognitive efforts. We argue that the formidable logical depth of mathematics has been the key to understand physical phenomena of large logical depth (and small Kolmogorov complexity), in a manner that our comparatively shallow human brains cannot match.

\end{enumerate} 

The very notion of {\it apparent complexity} has been previously studied in \cite{aaronson2014}, and was argued to be a likely transitional phase in closed systems whose initial state was of remarkably small entropy --- which means that our argument for the effectiveness of deep learning may have roots in the {\it second law of thermodynamics}. In particular, \cite{aaronson2014} presented four distinct measures of "apparent complexity", and proved that they all are  related. {\it Logical depth} was one of these four measures. This give us good reasons to believe that the current state of the universe has a remarkably large logical depth.

\section*{The Main Conjecture in Perspective}

Not surprisingly, several tentative explanations of the success of deep learning were recently proposed. Many share the same purpose, namely identifying a relevant space of functions that require either depth or exponentially large width \cite{hastad1990,braverman2011,bengio2011}. To simplify the theoretical analysis, these approaches usually rather focus on one particular kind of neural networks, e.g. ReLU in \cite{telgarsky2015,telgarsky2016}, ReLU, sigmoids and threshold in \cite{eldan2016} or logic gate circuits in \cite{yao1985}. Logic gate circuits can be regarded as specific kinds of neural networks whose inter-neural communications are bits. For instance, \cite{hastad1986} proved the existence of functions whose computations can be performed with a polynomial-size logic gate circuit of depth $k$, but which require exponentially many logic gates to be run on circuits of depth $k-1$.

A somewhat different approach has been taken by researchers from Google and Stanford \cite{raghu2016,poole2016}. Instead of searching for a specific space of functions that only deeper neural networks can compute, they determined typical properties that functions computed through deep neural networks have. To do so, they considered a Gaussian probability distributions over deep neural networks and showed that, for certain such probability distributions, a certain measure of the complexity of the functions computed by the random deep neural networks was increasing at a rate of the form $width^{\Theta(depth)}$ in expectation. In particular, the effect of depth is exponential, while width only acts in a polynomial way. The measures of the complexity of the functions differ in the two papers \cite{raghu2016,poole2016}, as each adapts its definitions to the neural networks under study. Yet both essentially boil down to some measure of nonlinearity.

One may also wonder how the results by \cite{raghu2016,poole2016} relate to our analysis. It is worth pointing out that there is a sense in which a large amount of nonlinearity is typical of large logical depth. Indeed, by opposition, the composition of linear operations remains a linear operation. As a result, an algorithm that combines a large number of linear operations is actually equivalent to a shorter algorithm that only computes their composition in a single operation. Therefore, the composition of linear operations does not increase logical depth. It is only the composition of nonlinear operations that may do so.

Now, it was already argued \cite{lin2016} that physical processes are fundamentally sequential, and hence any algorithm that attempts to understand physical data should be sequential as well. Our conjecture however differs from this claim. Indeed, we argue that data derived from sequential processes are not necessarily of large logical depth. As an example, \cite{aaronson2014} observe that the early universe and the far-end universe have low {\it apparent complexity}, even though they are derived from a large number of computational steps. In fact, \cite{aaronson2014} even argue that the large {\it apparent complexity} of our current universe is only a temporary phase which will vanish as entropy continues to increase. In any case, it is this fundamental computational property of our current universe, measured in terms of {\it non-parallelizable logical depth}, that our main conjecture relies on.

\section*{Corollaries and Further Steps}

An interesting corollary of our conjecture is that state-of-the-art predictive models whose predictions rely on a huge amount of unavoidable computation time (rather than on a huge amount of data) are unlikely to be superseded by current machine learning algorithms, as all current machine learning algorithms can be regarded as fast algorithms. Even the deepest neural networks currently being used hardly exceed a few hundred non-parallelized computation steps.


A major difficulty with training even deeper models would be the exploration (or learning) phase. Indeed, currently, a major bottleneck of machine learning is the huge computation power needed for learning (we, humans, have to spread our learning over decades). Models that would require larger computation time to make predictions would likely require larger computation time for learning as well. A solution to this may be to decompose the learning into different phases. This is actually how people learn to play chess, or other deep endeavors like mathematics or computer science. Instead of learning to play entire chess games repeatedly, chess players focus on independent  lower-logical-depth states of chess games, e.g. end games. 

An important further step could be to foresee the exact need for deeper learning, depending on the task at hand. This challenge appears for humans as well. Indeed, as argued by psychologists \cite{kahneman2011}, our brains seem to feature two thinking modes. One is fast, reactive and only partially reliable. It is a shallow learning part of our brains. The other is slow and more reliable. It is a deeper learning part of our brains. One important problem that our brains repeatedly need to solve is whether the slower part of our brain is needed to solve the task at hand. Such a problem will likely need to be solved for artificial intelligence as well.

\section*{Conclusion}

We proposed here a very first step towards understanding the success of 
deep learning. Basically, we argued that fundamental concepts in theoretical computer science, namely {\it Kolmogorov complexity}, {\it logical depth} and the {\it P vs NC} conjecture, could provide better insights into the nature of the data exploited by machine learning algorithms, and   better foresee which machine learning algorithms are most likely to succeed in this endeavor. In short, neural networks and specifically deep learning seems preferable over other approaches, mostly because neural networks allow for a facilitated exploration of a subspace of large-Kolmogorov-complexity algorithms, and deep learning better matches the large non-parallelizable logical depth of the current state of our universe. 

Formalizing our main conjecture goes through a rigorous, natural and exploitable definition of non-parallelizable logical depth, which is non-trival. It would then be interesting to mathematically prove that no shallow neural network can compute large non-parallelizable logical depth, but that deeper neural networks can. Moreover, determining the non-parallelizable logical depth of real data, as well as of specific (approximate) functions related to this data, e.g. chess playing or passing the Turing test, would be a major step towards a theoretical understanding of deep learning.

\bibliographystyle{abbrv} 
\bibliography{main.bib}  
%
\balancecolumns
\end{document}